\documentclass[10pt,twocolumn,letterpaper]{article}

\usepackage[pagenumbers]{cvpr} % To force page numbers, e.g. for an arXiv version

\usepackage[dvipsnames]{xcolor}

\definecolor{cvprblue}{rgb}{0.21,0.49,0.74}
\usepackage[pagebackref,breaklinks,colorlinks,citecolor=cvprblue]{hyperref}

\usepackage{float}

\def\paperID{91} % *** Enter the Paper ID here
\def\confName{3DV\xspace}
\def\confYear{2026\xspace}

\def\paperTitle{GeoSAM2: Unleashing the Power of SAM2 for 3D Part Segmentation}

\def\authorBlock{
    Ken Deng\textsuperscript{1,4*} \quad
    Yunhan Yang\textsuperscript{2*} \quad
    Jingxiang Sun\textsuperscript{3*} \\[0.5em]
    Xihui Liu\textsuperscript{2} \quad
    Yebin Liu\textsuperscript{3} \quad
    Ding Liang\textsuperscript{1} \quad
    Yan-Pei Cao\textsuperscript{1} \\[0.5em]

    \small
    \textsuperscript{1}VAST \quad
    \textsuperscript{2}The University of Hong Kong \quad
    \textsuperscript{3}Tsinghua University \quad
    \textsuperscript{4}Sun Yat-sen University \\[0.5em]

    \small
    \textsuperscript{*}Equal Contribution
}
\begin{document}

\title{\paperTitle}
\author{\authorBlock}
% \maketitle
\twocolumn[{
\renewcommand\twocolumn[1][]{#1}
\maketitle
\begin{center}
    % \vspace{-5em}
    \captionsetup{type=figure}
    % Adjust the scale or width to fit your document
    \includegraphics[width=\textwidth]{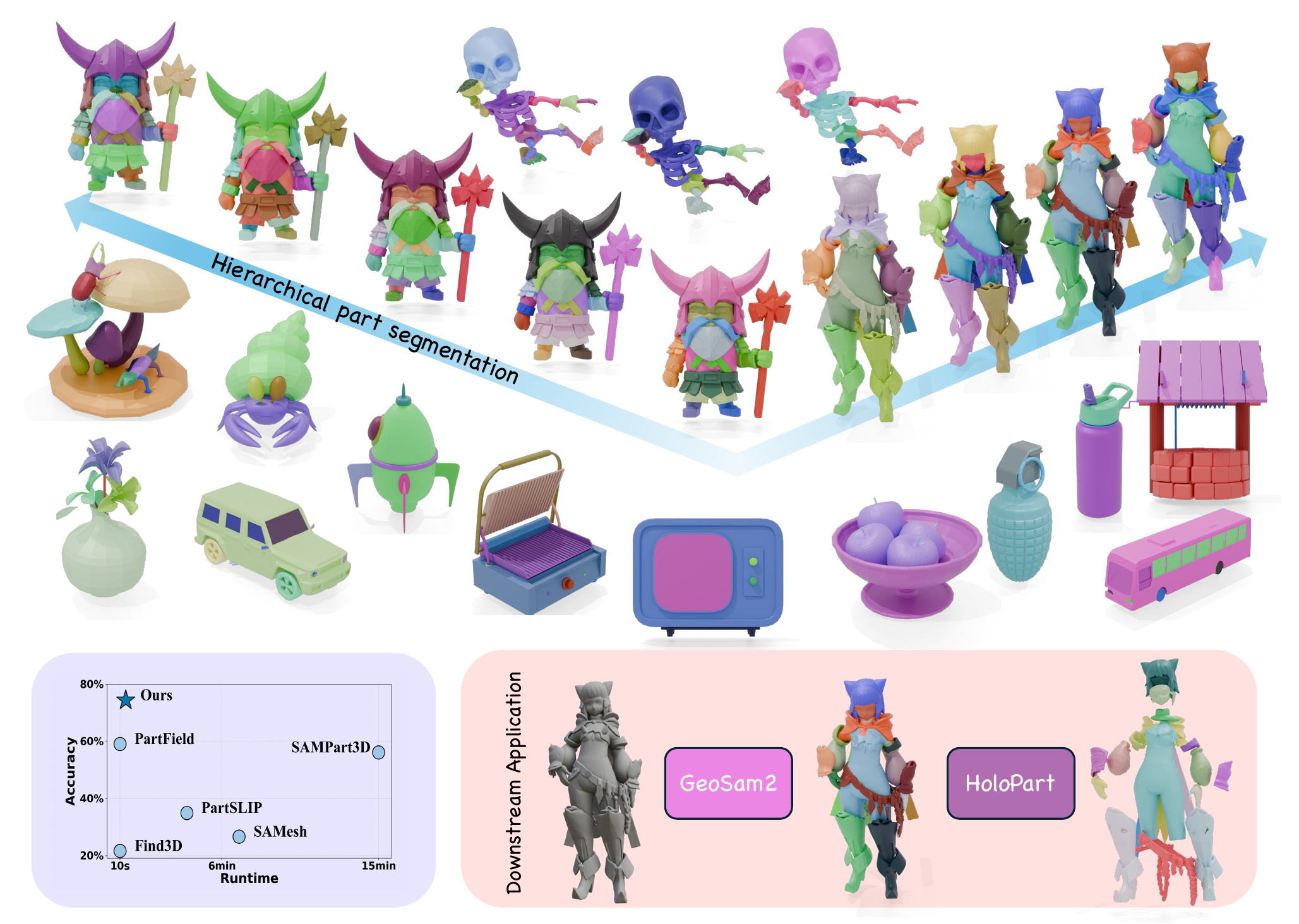} 
    \captionof{figure}{We introduce GeoSAM2, a scalpel-precision 3D shape segmentation method that enables arbitrary-detail segmentation through simple 2D prompt control. Demonstrating strong generalization on open-world objects, GeoSAM2 significantly outperforms baseline methods with minimal computational overhead. Its versatile capabilities open up broad applications, such as combining with 3D part completion models (e.g., HoloPart~\cite{holopart}) to achieve effective 3D part amodal segmentation.}
    \label{fig:teaser}
\end{center}
}]

\section{Abstract}
We introduce GeoSAM2, a prompt-controllable framework for 3D part segmentation that casts the task as multi-view 2D mask prediction. Given a textureless object, we render normal and point maps from predefined viewpoints and accept simple 2D prompts—clicks or boxes—to guide part selection. These prompts are processed by a shared SAM2 backbone augmented with LoRA and residual geometry fusion, enabling view-specific reasoning while preserving pretrained priors. The predicted masks are back-projected to the object, aggregated across views.
Our method enables fine-grained, part-specific control without requiring text prompts, per-shape optimization, or full 3D labels. In contrast to global clustering or scale-based methods, prompts are explicit, spatially grounded, and interpretable. We achieve state-of-the-art class-agnostic performance on PartObjaverse-Tiny and PartNetE, outperforming both slow optimization-based pipelines and fast but coarse feedforward approaches. Our results highlight a new paradigm: aligning the paradigm of 3D segmentation with SAM2, leveraging interactive 2D inputs to unlock controllability and precision in object-level part understanding.

% aligning interactive 2D inputs with 3D segmentation unlocks controllability and precision in mesh-level part understanding.

\section{Introduction}

3D part segmentation is a crucial and challenging task in 3D perception, playing a vital role in downstream applications such as robotic manipulation~\cite{van2024open}, 3D generation~\cite{holopart}, and interactive editing~\cite{tertikas2023generating}. Despite impressive results on curated benchmarks, fully supervised 3D methods are limited by scarce annotations: acquiring detailed part labels for 3D models is time-consuming and labor-intensive~\cite{li2018pointcnn}.

This bottleneck has motivated zero-shot and weakly supervised approaches. In particular, recent works have leveraged powerful 2D vision foundation models to project segmentation cues into 3D~\cite{yang2023sam3d, find3d, tang2024segment}. For example, SAMPart3D~\cite{yang2023sam3d} casts part segmentation as a multi-granularity clustering problem by distilling 3D features conditioned on a continuous “scale” parameter to yield fine or coarse parts. While this enables zero-shot operation without text prompts, the scale knob is inherently unintuitive: varying the scale often produces unpredictable splits without semantic grounding. Moreover, SAMPart3D requires per-shape MLP fitting to realize the scale-conditioned grouping, which is computationally expensive (minutes per object) and incompatible with real-time usage. As a result, user control is coarse and inflexible.
% and the segmentation lacks semantic precision.

Other methods aim to leverage either 2D mask proposals or feedforward field representations. SAMesh~\cite{tang2024segment} lifts 2D masks from SAM2 to 3D by rendering a mesh from multiple viewpoints and aggregating the per-view predictions. This avoids text prompts and leverages SAM2’s open-vocabulary capabilities, but it lacks any mechanism for part-specific queries. The pipeline is slow because its post-processing involves iterative optimization on mesh, which can take several minutes per mesh. In contrast, PartField learns a continuous part-feature field in a single feedforward pass, enabling fast inference and view-consistent segmentation. However, it also suffers from limited controllability: users can only coarsely adjust granularity via fixed cluster counts, and it assumes all object regions are labelable. The global and coarse control hampers accurate mask prediction, since scale alone is insufficient to capture user intent.

% On datasets with incomplete annotations (e.g., PartNetE), this leads to arbitrary or incorrect segmentation.

In summary, existing methods exhibit key limitations: control over segmentation is indirect and coarse-grained; and most pipelines are either fast but inflexible (e.g., PartField) or expressive but slow (e.g., SAMPart3D, SAMesh). Moreover, none fully aligns 2D interaction with 3D part outcomes, leaving a gap in accurate 3D segmentation.

To address these gaps, we reformulate the problem of prompt‐controllable 3D part segmentation for textureless object as a multi‐view 2D mask task. Given a 3D model $\mathcal{M}$, we render a small set of normal and point maps from canonical viewpoints and accept simple 2D prompts (clicks or boxes) on the user-selected view to indicate the target part. These view–prompt pairs are processed by a shared SAM2 backbone—extended via Low-Rank Adaptation in every transformer block—to inject geometric structure while leaving the majority of the network frozen. At each Feature Pyramid Networks (FPN) level, zero-initialized residual convolutions fuse normal- and point-map features, preserving pretrained RGB feature statistics and gradually learning cross-modal associations. To enforce cross-view consistency, we retain embeddings from all views in a full-view memory bank (bootstrapped by duplicating the first frame), then back-project per-view masks onto $\mathcal{M}$. A lightweight post-processing step removes small components and smooths labels via k-NN voting. Because our 3D labels are exactly aligned with the 2D masks, we not only enable intuitive, region-specific control—unlike PartField’s global scale knob—but also deliver higher accuracy on both fully and partially labeled datasets (e.g., PartNetE) without forcing every face to carry a label.

Experimental results demonstrate that our model achieves SOTA class-agnostic segmentation performance on the PartObjaverse-Tiny and PartNetE benchmarks, significantly outperforming existing methods. We summarize our main contributions below:
\begin{itemize}
    \item Reformulate 3D part segmentation as an interactive 2D prompt-based multi-view segmentation task that unleashes the power of 2D segmentation foundation models for fine-grained, part-specific control over 3D surfaces.
     % (e.g., SAM2)

    \item Design a geometry-aware encoder with LoRA-based parameter-efficient tuning and residual fusion of normal and point-map features, allowing effective adaptation to textureless 3D inputs while preserving pretrained  priors. 
    % feature

    \item Achieve SOTA class-agnostic segmentation performance on PartObjaverse-Tiny and PartNetE benchmarks, demonstrating high segmentation accuracy and prompt-aligned 3D labeling.
    % —even under partially annotated supervision.

\end{itemize}

\section{Related Work}

\noindent\textbf{Lifting 2D Foundation Models for 3D Segmentation}. Recent approaches leverage the powerful representation capabilities of pre-trained 2D foundation models, such as CLIP~\citep{radford2021clip}, GLIP~\citep{li2022grounded}, SAM~\citep{kirillov2023segment}, and DINOv2~\citep{oquab2023dinov2}, to address the scarcity of annotated 3D data. Typically, these methods render 3D model from multiple views, apply a 2D foundation model to each view, and then aggregate the results into 3D predictions~\citep{liu2023sanerfhq,zhou2024serffinegrainedinteractive3d,cen2024segment3dradiancefields}. SAM3D~\citep{yang2023sam3d} and SAMPro3D~\citep{xu2023sampro3d} extend SAM by fusing RGB-D view segmentations into coherent 3D masks. Similarly, PartSLIP~\citep{liu2023partslip} and its enhanced version PartSLIP++\citep{zhou2023partslip++} first use GLIP and SAM to propose and refine part masks in multiple views before consolidating these into 3D part labels. Another paradigm, exemplified by Segment3D\citep{huang2023segment3d}, SAL~\citep{ovsep2024better}, and Point-SAM~\citep{zhou2024point}, employs SAM-generated pseudo-labels to train dedicated 3D segmentation models directly. Alternatively, methods like OpenScene~\citep{peng2023openscene} project CLIP features from multiple views into the 3D space for zero-shot scene parsing. Recently, Garosi et al.~\citep{garosi20253d} proposed COPS, which extracts DINOv2 features from rendered views and incorporates geometric-aware feature aggregation to enhance spatial and semantic coherence.

However, these lifting methods face several limitations. Most depend heavily on elaborate prompt engineering to align 2D models with the 3D task~\citep{zhu2023pointclipv2,garosi20253d}. Furthermore, aggregation across views typically relies on heuristic rules or bounding-box constraints, resulting in noisy and computationally expensive processes~\citep{liu2023partslip,zhou2023partslip++,yang2023sam3d}. Crucially, these methods rarely leverage intrinsic geometric structure, leading to inconsistencies, particularly for occluded parts or textureless regions. Detection-based methods (e.g., PartSLIP~\citep{liu2023partslip}) also inherently neglect internal structures that are not externally visible. Addressing these challenges, recent efforts advocate incorporating geometric priors explicitly to achieve more robust and geometrically consistent 3D segmentation outcomes~\citep{garosi20253d}.

\noindent\textbf{Conventional 3D Part Segmentation.} Traditional 3D part segmentation relies on either supervised learning or geometric heuristics. Deep networks like PointNet++\cite{qi2017pointnet}, MeshCNN\cite{Hanocka_2019}, Point Transformer~\cite{zhao2021point}, and Laplacian Mesh Transformer~\cite{10.1007/978-3-031-19818-2_31} learn to predict parts given labeled meshes. However, existing benchmarks (e.g., PartNet~\cite{mo2019partnet}, COSEG~\cite{wang2023coseg}, Princeton Mesh Segmentation~\cite{Chen2009}) cover only limited object classes, so label scarcity remains a bottleneck. Geometry-based methods, (e.g., Shape Diameter Function(SDF)~\cite{Shapira2008ConsistentMP,Roy_2023}), segment a mesh using local thickness cues. These classic algorithms often fail on complex or fine-grained parts and require careful parameter tuning.

\noindent\textbf{Zero-Shot 3D Part Segmentation.} To reduce reliance on 3D labels, recent works exploit pretrained 2D models via multi-view rendering or knowledge distillation. A common strategy is to lift 2D segmentation into 3D: SATR~\cite{abdelreheem2023satrzeroshotsemanticsegmentation} employs a 2D detector conditioned on text prompts to label semantic parts of a mesh, and MeshSegmenter~\cite{zhong2024meshsegmenter} synthesizes textures so that SAM can segment the mesh by color cues. These methods produce semantic masks for user-specified part categories but do not automatically discover all parts of an object. Other approaches transfer 2D knowledge to 3D via learning: PartSLIP~\cite{liu2023partslip} leverages the GLIP detector~\cite{li2022glip} on projected views to detect part instances, and PartSLIP++\cite{zhou2023partslip++} refines these detections with SAM-generated masks. ZeroPS\cite{xue2023zerops} introduces a two-stage pipeline that aligns multi-view geometry with vision-language prompts to segment parts. PartDistill~\cite{umam2023partdistill} trains a 3D network by distilling 2D segmentation masks into the model, and SAMPart3D~\cite{yang2024sampart3d} extends this idea by pretraining on large 3D datasets with 2D supervision. These methods achieve zero-shot semantic part segmentation via 2D-to-3D transfer, but they still rely on predefined label sets or prompt engineering.
\section{Preliminary}
\paragraph{Segment Anything Model 2 (SAM2).}
SAM2 extends the original \emph{Segment Anything Model} from images to \emph{promptable} video segmentation by attaching a lightweight, memory-augmented transformer.  
For each frame \(I_t\) it extracts image features \(\mathbf{f}_t=\mathcal{E}_{\text{img}}(I_t)\) with a hierarchical ViT encoder, and obtains a prompt embedding \(\mathbf{p}_t=\mathcal{E}_{\text{prompt}}(\text{prompt})\).  
A cross-attention module conditions the frame on the memory bank of past object tokens, yielding
\(\tilde{\mathbf{f}}_t=\operatorname{MemAtt}\!\bigl(\mathbf{f}_t,\mathcal{M}_{t-1}\bigr)\).  
The mask decoder then predicts the object mask
\(\hat{M}_t=\mathcal{D}\!\bigl(\tilde{\mathbf{f}}_t,\mathbf{p}_t\bigr)\).  
Finally, the memory encoder compresses \((\hat{M}_t,\mathbf{f}_t)\) into a token
\(\mathbf{m}_t=\mathcal{E}_{\text{mem}}\!\bigl(\hat{M}_t,\mathbf{f}_t\bigr)\),
which is enqueued into \(\mathcal{M}_t\) for subsequent frames.  
This four-stage loop—image encoder, prompt + mask decoder, memory attention, and memory encoder—produces temporally consistent, real-time masks across a video sequence.

% an entire 
\section{Method}

\setlength{\abovedisplayskip}{3pt} % 公式上方间距
\setlength{\belowdisplayskip}{3pt} % 公式下方间距

\begin{figure*}[t]
  \centering
  \includegraphics[width=\linewidth]{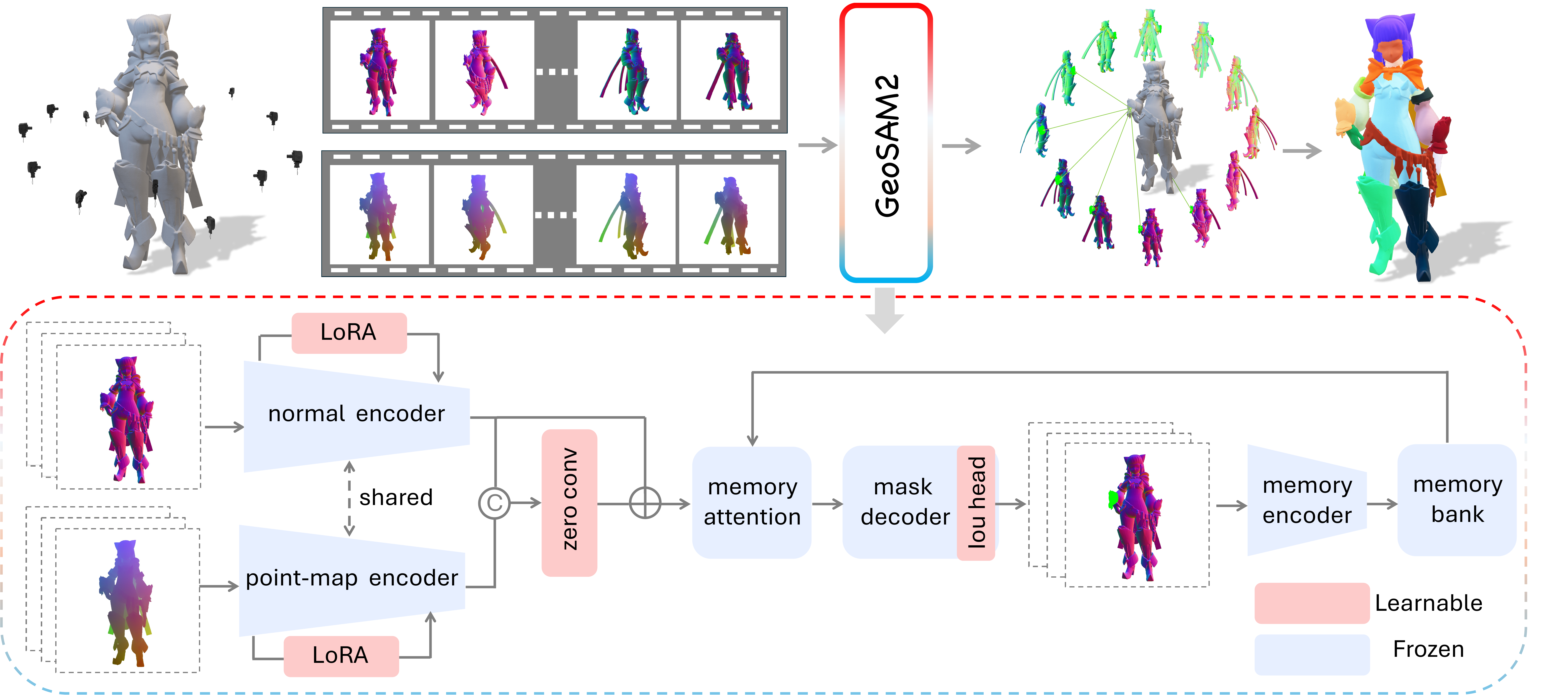}
  \caption{Our pipeline renders 12-view normal and point maps of an object, arranged into an inverse-clockwise video sequence. Users can annotate any frame with 2D prompts, which serves as the video's starting frame. GeoSAM2 processes these inputs by first encoding each frame's normal and point maps using pretrained (frozen) image encoders fine-tuned with LoRA, fusing their features, and decoding masks. The per-frame 2D masks are then projected onto a 3D point cloud using camera poses, with visibility-aware voting assigning consistent labels across views.}
  \label{fig:pipeline}
\end{figure*}

% Key adaptations include fine-tuning the mask decoder's IoU head to match dataset-specific annotation scales, ensuring robust 3D segmentation.

We address the problem of part segmentation for textureless 3D model by formulating it as a multi-view segmentation task. Given a 3D model $\mathcal{M}$, we render a set of normal maps $\{I_i\}_{i=1}^N$ and point maps $\{\Pi_i\}_{i=1}^N$ from predefined camera poses $\{P_i\}_{i=1}^N$, capturing geometric details from multiple viewpoints. Specifically, we arrange these multi-view renderings into a 12-frame video by sequencing the viewpoints counterclockwise around the mesh. Each image $I_i$ is processed by a vision model $\mathcal{F}$ that integrates geometric information to predict per-pixel part masks $\{M_i\}_{i=1}^N$. To ensure consistency across views, we employ a memory mechanism $\mathcal{T}$ that tracks and associates masks $M_i$ across different views, maintaining coherent part labels. The aggregated multi-view masks are then projected back onto the model surface $\mathcal{M}$ using the inverse of the rendering transformations, resulting in an initial 3D part segmentation. Finally, a post-processing step $\mathcal{P}$ refines the segmentation by smoothing the boundaries and correcting inconsistencies. The overall pipeline is shown in Figure~\ref{fig:pipeline}.

We first describe our data strategy, which constructs semantically meaningful part annotations via mesh-based decomposition (Section~\ref{data_curation}). Next, we introduce a geometry-aware encoder with low-rank adaptation, enabling efficient transfer from RGB to geometric modalities (Section~\ref{lora}). We then propose a residual multi-modal fusion scheme to combine normal and point-map features progressively (Section~\ref{featfusion}). Finally, we redesign the memory mechanism for multi-view consistency and apply mesh-level post-processing to refine part labels (Section~\ref{memorybank}-\ref{postprocess}).

\subsection{Data Curation}
\label{data_curation}
% For each object, we first decompose it into parts according to the connectivity of the mesh, then compose them according to regulars: (1) Each part should have obvious semantic information, i.e., can be well described by language. (2) Every part should be distributed at similar locations in space. (3) Different parts should have similar volumes. (4) An object shouldn't be split into more than 15 parts. For those objects that have too many parts or only consist of one part, they will be discarded. More details can be found in supplementary material.

For each object, we first decompose it into parts according to the connectivity of the mesh. For those objects decomposed to much parts (higher than 15), we recompose them to reduce part number by combining similar parts. For parts that are far apart in space (connected by thin faces), we split them into different parts to ensure that each part is physically continuous in space. More details can be found in the supplementary material.

\subsection{Geometry-Aware Encoding with Low-Rank Adaptation}
\label{lora}
\paragraph{LoRA based Geometry-Aware Encoders.} Although SAM2 is trained on RGB domain and performs well in multi-scale mask prediction, applying it to multi-view segmentation of textureless 3D objects poses significant challenges. The core issue lies in the reliance of SAM2 on RGB-based appearance cues, which are absent in textureless renderings such as normal maps. Additionally, unlike dense temporal videos, sparse multi-view renderings exhibit large viewpoint shifts and complex occlusions, making it difficult to establish accurate correspondences across views. 

% Furthermore, SAM2’s backbone $\mathcal{F}$—a MAE-based vision encoder—lacks the semantic abstraction capacity required to group geometrically similar but visually indistinct regions.

To address these limitations, we inject geometric structure directly into the model via geometry-aware features derived from normal and point maps. Given a normal map $I_i$ rendered from camera pose $P_i$, we pair it with a corresponding point map $\Pi_i$, which is obtained by back-projecting the depth map into 3D using the camera parameters. This gives each pixel a 3D coordinate in the world frame:
\[
\mathbf{x}_{i}(u, v) = D_i(u, v) \cdot K^{-1} [u, v, 1]^T R_i^{-1} - R_i^{-1} \mathbf{t}_i,
\]
where $D_i$ is the depth map rendered from $\mathcal{M}$, $(u,v)$ is a pixel coordinate, and $(R_i, \mathbf{t}_i)$ is the camera extrinsic matrix from $P_i$. The resulting point map $\Pi_i$ encodes view-consistent spatial structure that helps resolve ambiguity and enables correspondence reasoning between views.

 % surface regions

However, directly training the full model $\mathcal{F}$ to adapt from RGB to geometric modalities (normal and point maps) is computationally inefficient. Inspired by recent advances in parameter-efficient adaptation, we adopt Low-Rank Adaptation (LoRA) to update only a small subspace of $\mathcal{F}$. Specifically, we instantiate two independent LoRA branches for the normal map and point map, repectively. For any linear projection layer in $\mathcal{F}$ with weight $W_0 \in \mathbb{R}^{m \times n}$, we introduce trainable matrices $A \in \mathbb{R}^{m \times r}$ and $B \in \mathbb{R}^{r \times n}$ such that the adapted weight becomes:
\[
W = W_0 + AB, \quad r \ll \min(m,n).
\]
Thus, for an input vector $\mathbf{f}$, the output is:
\[
W \mathbf{f} = W_0 \mathbf{f} + A (B \mathbf{f}),
\]
while $W_0$ remains frozen. We find that injecting LoRA layer per transformer block suffices to steer $\mathcal{F}$ towards effectively processing geometry-derived modalities.

\paragraph{Residual Fusion of Normal and Point-Map Features.}
\label{featfusion}
Normal-map features are more closely aligned with RGB textures and thus better match the pretrained statistics of SAM2's backbone. In contrast, point-map features carry strong geometric cues but may introduce distributional shift if fused naively. To balance these modalities, we propose a residual fusion strategy that initializes conservatively and adapts progressively.

At each resolution of the feature pyramid, we are given a pair of aligned features: the normal-map feature $G_i \in \mathbb{R}^{H \times W \times C}$ and the point-map feature $P_i \in \mathbb{R}^{H \times W \times C}$. These are first concatenated along the channel axis:
\[
X_i = [\,G_i \,\Vert\, P_i\,] \in \mathbb{R}^{H \times W \times 2C},
\]
and passed through a $3 \times 3$ convolutional layer whose weights are initialized to zero:
\[
Y_i = \operatorname{Conv}_{3 \times 3}(X_i; W = 0) \in \mathbb{R}^{H \times W \times C}.
\]
We then apply a residual connection by adding the result back to the original normal feature:
\[
\hat{G}_i = G_i + Y_i.
\]

Because the convolution starts with zero weights, the initial output $Y_i$ is exactly zero, and the network relies solely on the normal-map features at the beginning of training. This prevents sudden shifts in feature distribution and allows gradients to gradually shape the contribution of the point-map branch. The same fusion strategy is applied independently at each resolution level in the FPN.

% This design preserves pretrained feature statistics and allows the model to progressively learn cross-modal associations. As a result, the network becomes better at segmenting semantically related parts that are spatially distant or geometrically ambiguous—especially under sparse views and in the absence of strong texture cues.

\subsection{Redesigning Memory Mechanism for Viewpoint-Aware Segmentation}
\label{memorybank}
SAM2 was originally designed for video segmentation, employing a fixed-size memory bank to retain features from recent frames and maintain temporal consistency. However, this design assumes redundant information and smooth transitions between adjacent frames—conditions that do not hold in multi-view 3D segmentation. In multi-view settings, each input viewpoint carries unique spatial cues and various informative geometric features. Consequently, discarding earlier views using SAM2's FIFO memory mechanism can lead to irreversible information loss and reduced segmentation accuracy and consistency.

\paragraph{Multi-View Memory Retention.} To overcome this limitation, we reorganize the memory mechanism by retaining features from all views throughout the sequence. Specifically, we render each object from 12 canonical views evenly spaced in azimuth and distributed across three elevation angles ($25^\circ$, $0^\circ$, and $-25^\circ$). This arrangement ensures comprehensive geometric coverage and allows the model to effectively reason about occlusions and part boundaries by referencing the entire set of previously encountered views. Retaining the view memory preserves critical cues from earlier perspectives and enhances cross-view consistency, providing a stable reference throughout the entire process.

\paragraph{Memory Bootstrapping via Frame Repetition.} Even with memory retention, we observe that SAM2 produces poor segmentation quality on the first frame of a sequence since the memory bank initially starts empty and the model can rely only on sparse prompts such as points or bounding boxes. Interestingly, segmentation improves markedly once memory is populated, indicating that the memory bank functions not only as a temporal consistency mechanism, but also as an implicit form of guidance. Inspired by this insight, we propose a simple yet effective bootstrapping strategy: duplicating the first frame before processing begins. This duplication provides an immediate and meaningful memory prior, significantly improving initial segmentation quality and leading to sharper, more coherent masks across subsequent views.

\subsection{Post-Processing Refinement}
\label{postprocess}
Through the previous steps, we can already obtain high-quality multi-view masks and 3D masks can be easily obtained through backprojecting multi-view masks to 3D model. For those mesh-form 3D model, masks can be post-processed by leveraging the connectivity of the mesh. Inspired by SAMesh, we propose a novel post-processing: (1) we first remove those components that have small area less than $A_{\text{mesh}} = P A_{\text{mesh}} \cdot N_{\text{faces}}$ (e.g., $P A_{\text{mesh}}$ = 0.01), (2) then smooth the mesh labels. Iteratively smoothing the face labels by connectivity is optional because it is time-consuming, then voting for the rest faces by K nearest neighbour to guarantee every face has label.

% To obtain 3D mask, we first backproject the label from pixel to face. However, some areas are invisible in multi-view masks, and back-projection of areas with drastic depth changes under multi-views will inevitably bring noise. Inspired by SAMesh, (1) we first remove those components that have small area less than $A_{\text{mesh}} = P A_{\text{mesh}} \cdot N_{\text{faces}}$ (e.g., $P A_{\text{mesh}}$ = 0.01), (2) then smooth the mesh labels. Iteratively smoothing the face labels by connectivity is optional because it is time-consuming, then voting for the rest faces by K nearest neighbour to guarantee every face has label.

\section{Experiments}

\setlength{\abovecaptionskip}{1pt} % 设置为0以缩短距离

\begin{figure*}
  \centering
  \includegraphics[width=\linewidth]{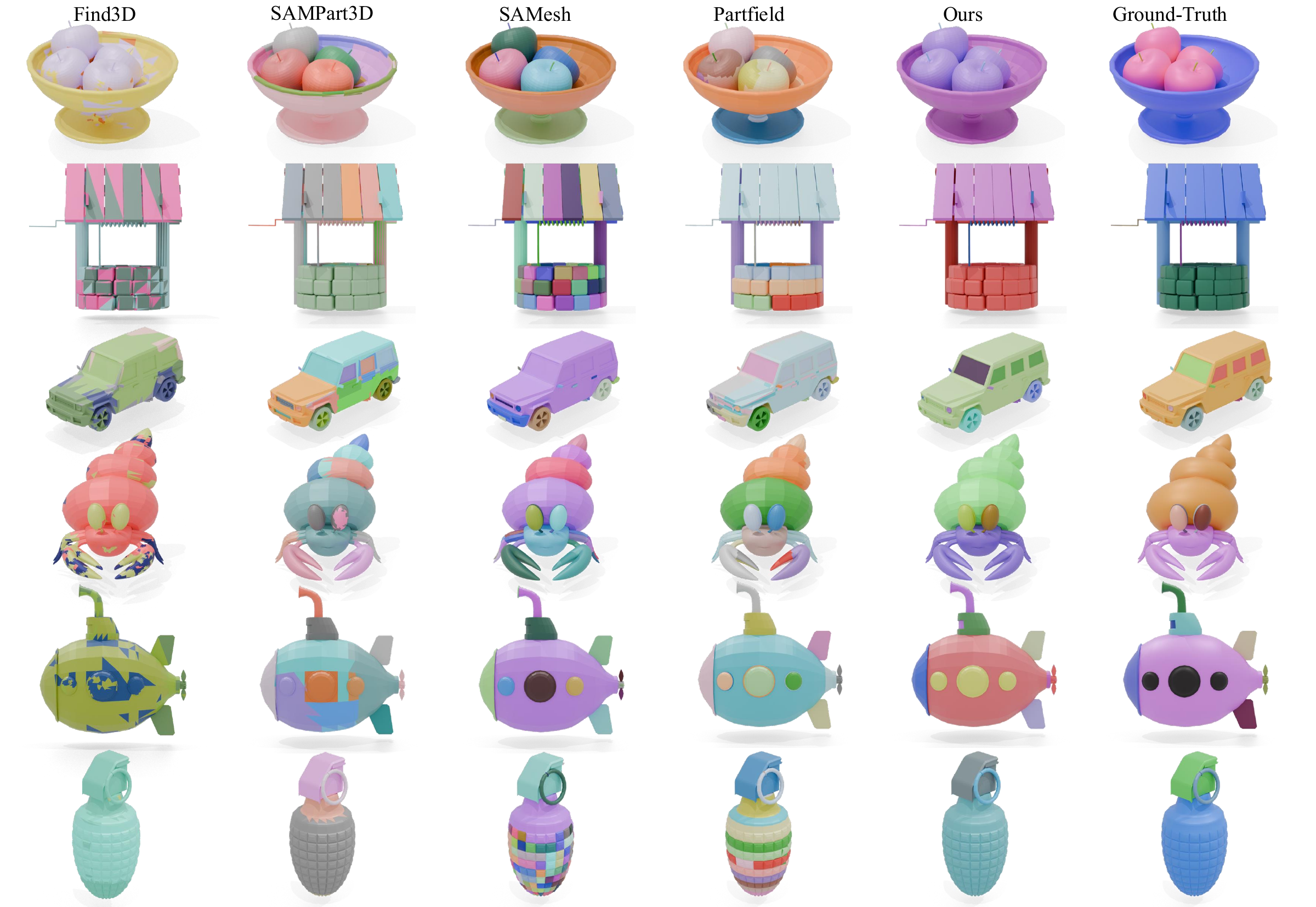}
\caption{\footnotesize Qualitative comparison of class-agnostic segmentation on the PartObjaverse-Tiny dataset~\cite{yang2024sampart3d}. The baselines include Find3D~\cite{find3d}, SAMesh~\cite{tang2025segmentmesh}, and SAMPart3D~\cite{yang2024sampart3d}, PartField~\cite{liu2025partfield}. Each color represents a different part. More results can be found in supplementary material.}
  \label{fig:comparison}
  \vspace{-5pt}
\end{figure*}

\subsection{Implementation Details}
To keep the strong prior of SAM2, we froze most parameters, mainly training two LoRA modules with rank 4 to all Q, K, V attention layers in the network and the feature fusion block, which consists of 3 convolution layers with a kernel size of 3. To encourage SAM2 to predict masks that is close to the scale of the dataset mask, we also train the iou prediction head. We fine-tuned SAM2, base plus version, 50 epochs on a self-annotated dataset containing about 4700 objects using 8 A800 with batch size of 8, learning rate of 5e-5 and we adopt the same loss as used in SAM2. To ensure stable training, we zero init the convolution layers of feature fusion block. 

% \begin{figure*}
%   \centering
%   \includegraphics[width=\linewidth]{Images/vis_shadow_paper_2.pdf}
% \caption{\footnotesize Qualitative comparison of class-agnostic segmentation on the PartObjaverse-Tiny dataset~\cite{yang2024sampart3d}. The baselines include Find3D~\cite{find3d}, SAMesh~\cite{tang2025segmentmesh}, and SAMPart3D~\cite{yang2024sampart3d}, PartField~\cite{liu2025partfield}. Each color represents a different part. More results can be found in supplementary material.}
%   \label{fig:comparison}
%   \vspace{-5pt}
% \end{figure*}

% \begin{figure}
%   \centering
%   \includegraphics[
%     width=\columnwidth,
%   ]{Images/hierarchy.pdf}
%   \caption{Visualization of hierarchical segmentation results of generative models. 3D models are segmented hierarchically, starting with the finest level of detail at the top, and then proceeding with varying granularities of segmentation for the hands, legs, and head. More results can be found in supplementary material.}
%   \label{fig:hierarchy}
%     \vspace{-2pt}
% \end{figure}

\begin{table}[t]
  \centering
  \captionsetup{skip=2pt}
  \caption{\footnotesize Quantitative evaluation of class‑agnostic part segmentation on PartObjaverse-Tiny~\cite{yang2024sampart3d}. Instance‑level labels; metric: mean IoU.}
  \label{tab:compare_pot}
   \resizebox{\linewidth}{!}{%
  \begin{tabular}{l|cccccccc|c}
    \toprule
    \textbf{Method} & Human & Animals & Daily & Build.\& & Transp. & Plants & Food & Elec. & Avg. \\
    \midrule
    Find3D~\cite{find3d}     & 26.17 & 23.99 & 22.67 & 16.03 & 14.11 & 21.77 & 25.71 & 19.83 & 21.28  \\
    SAMPart3D~\cite{yang2024sampart3d}  & 55.03 & 57.98 & 49.17 & 40.36 & 47.38 & 62.14 & 64.59 & 51.15 & 53.47  \\
    SAMesh~\cite{tang2024segmentmeshzeroshotmesh} & 66.03 & 60.89 & 56.53 & 41.03 & 46.89 & 65.12 & 60.56 & 57.81 & 56.86  \\
    PartField~\cite{liu2025partfield}                       & 80.85 & 83.43 & 77.83 & 69.66 & 73.85 & 80.21 & \textbf{85.27} & 82.30 & 79.18  \\
    \textbf{Ours}                       & \textbf{88.99} & \textbf{91.50} & \textbf{86.04} & \textbf{74.57} & \textbf{77.60} & \textbf{88.92} & 82.72 & \textbf{84.95} & \textbf{84.06}  \\
    \bottomrule
  \end{tabular}}
\end{table}

\begin{table}[t]
  \centering
  \captionsetup{skip=2pt}
  \caption{\footnotesize Quantitative evaluation of class‑agnostic part segmentation on PartNetE~\cite{liu2023partsliplowshotsegmentation3d}. Instance‑level labels; (mean IoU, \%).}
  \label{tab:compare_partnete}
  \resizebox{\linewidth}{!}{%
  \begin{tabular}{l|ccccc|cc}
    \toprule
    \textbf{Method} & Electro. \& Comput. & Home Appl. & Kitchen \& Food & Furnit. \& Househo. & Tools, Offi., \& Misc. & Avg. & Run‑time \\
    \midrule
    
    Find3D~\cite{find3d}               & 14.58 & 22.66 & 24.89 & 21.72 & 29.11 & 21.69 & \textbf{$\sim$10s} \\
    SAMPart3D~\cite{yang2024sampart3d} & 43.86 & 40.57 & 65.18 & 57.50 & 65.85 & 56.17 & $\sim$15min\\
    SAMesh~\cite{tang2024segmentmeshzeroshotmesh} & 23.96 & 26.73 & 31.50 & 22.82 & 33.49 & 26.66 & $\sim$7min\\
    PartField~\cite{liu2025partfield}  & 43.70 & 52.37 & 69.85 & 60.22 & 66.33 & 59.10 & \textbf{$\sim$10s}\\
    \textbf{Ours}                      & \textbf{69.93} & \textbf{71.23} & \textbf{79.73} & \textbf{71.97} & \textbf{79.41} & \textbf{74.42} & $\sim$30s \\
    \bottomrule
  \end{tabular}}
\end{table}

\begin{table}[t]
  \centering
  \captionsetup{skip=2pt}
  \caption{\footnotesize Ablation study (mean IoU, \%). Datasets as rows.}
  \label{tab:comparison_abla}
  \resizebox{\linewidth}{!}{%
  \begin{tabular}{l|cccc}
    \toprule
    \textbf{Dataset} & \textbf{Vanilla SAM2} & \textbf{w/o point map} & \textbf{w/o feature fusion} & \textbf{Ours} \\
    \midrule
    PartObjaverse-Tiny & 62.59 & 75.56 & 81.39 & 84.06 \\
    PartNetE           & 66.55 & 71.26 & 72.25 & 74.42 \\
    \bottomrule
  \end{tabular}
  }
\end{table}

\subsection{Comparision}

% \begin{figure*}
%   \centering
%   \includegraphics[width=\linewidth]{Images/objaverse_comparison_3.pdf}
% \caption{\footnotesize Qualitative comparison of class-agnostic segmentation on the PartObjaverse-Tiny dataset~\cite{yang2024sampart3d}. The baselines include Find3D~\cite{find3d}, SAMesh~\cite{tang2025segmentmesh}, and SAMpart3D~\cite{yang2024sampart3d}, PartField~\cite{liu2025partfield}. Each color represents a different part.}
%   \label{fig:comparison}
% \end{figure*}

\noindent \paragraph{Evaluation Datasets} We conduct evaluations on the PartObjaverse-Tiny and PartNetE datasets, following~\cite{liu2025partfield} in open-world 3D part segmentation. PartObjaverse-Tiny comprises 200 meshes across diverse object categories, each annotated with human-labeled part segmentations. PartNetE, derived from the PartNet-Mobility dataset, includes 1,906 point clouds from 45 object categories and provides annotations for movable parts. The category group mapping can be found in supplementary material.

\noindent \paragraph{Metric} In line with previous works, we adopt the class-agnostic mean Intersection over Union (mIoU) as our evaluation metric. For each ground-truth part, we compute the IoU against all predicted parts and assign the highest IoU as its score. The mIoU is then obtained by averaging these scores across all ground-truth parts.

\noindent \paragraph{Baselines} We compare GeoSAM2 with 4 recent state-of-the-art methods for open-world 3D part segmentation, all of which have been introduced in the latest literature. Find3D is a text-input part segmentation approach that leverages a feedforward model to align 3D feature spaces with text feature spaces, SAMPart3D employs 3D pretraining to distill multi-view DINOv2 features into a 3D encoder, but still requires per-shape finetuning based on multi-view SAM predictions. SAMesh applies a carefully designed community detection algorithm to lift multi-view predictions into the 3D domain. PartField operates directly on point clouds and predicts a feature field using a triplane representation, from which 3D part segmentations are obtained via clustering.

For all methods, we follow their publicly released implementations and evaluate them on both datasets. For text-based approaches, we utilize the annotated part labels from the dataset. For approaches that produce multi-scale part segmentations (e.g., SAMPart3D and PartField), we generatie 20 segmentation results with varying scales or cluster numbers. For our method, due to the object orientation uncertainty, we select 4 orthographics views as input respectively and adopt the highest mIoU. Considering that a view can't contain all parts of a 3D model, we also automatically segments the unlabeled area of the view which lies opposite the input view (Users can also replace this step).  

% Because PartNetE purely consists of point cloud and some point clouds are not fully marked, we didn't use post-processing and skipped the unlabeled points.

% For the post-processing, due to the irregular of mesh in PartObjaverse-Tiny, we select 3 different values of $A_{\text{mesh}}$, to be specific, 0.01, 0.02, 0.035.

\noindent \paragraph{Results} 
The evaluation results on PartObjaverse-Tiny are shown in Table~\ref{tab:compare_pot} and Figure~\ref{fig:comparison} presents comparative visual results. The mIoU of the baselines is the same as reported in~\cite{liu2025partfield}, as we adopt the same experimental setup. We observe that Find3D, a text-based method, underperforms compared to class-agnostic methods (Average mIoU: 21.28), which indicates that accurate 3D part detection with open-world semantics remains challenging. Although SAMesh can produce acceptable results (Average: 56.86), it always tends to segment a mesh into too fine-grained segments, and lacks controllability. SAMPart3D (Average: 53.47) and PartField (Average: 79.18) control the segmentation fineness via different scales, adjusting global granularity. Although global control can reduce prompt operations, lacking localized control signals makes fine-grained control inaccessible. In contrast, our method can control every part of a 3D model at different scales---same as SAM2, our method supports 2D prompts on the target view, and the paradigm empowers users to achieve pixel-level precision in segmentation outcomes, without laborious operation. 

% While our method achieves significant improvements over baselines on PartObjaverse-Tiny dataset, this benchmark inherently favors post-processing techniques leveraging mesh connectivity priors, partially obscuring our model’s intrinsic geometric reasoning capability. In contrast, PartNetE's exclusive use of raw point clouds (1) eliminates post-processing biases from mesh connectivity priors and (2) provides more unambiguous evaluation of segmentation capability. 

The evaluation results on PartNetE are shown in Table~\ref{tab:compare_partnete}. SAMesh and PartField exhibit 20\%-30\% mIoU collapse on PartNetE compared to its PartObjaverse-Tiny performance, directly exposing its dependency on mesh connectivity priors. In contrast, the performance on PartNetE of Find3D and SAMPart3D basically has not changed because their post-processing exclusively relies on k-nearest neighbors. While our method experiences a reduction to some extent when deprived of mesh connectivity priors, it maintains a significant lead over all baseline approaches.

% In terms of runtime, our method takes ~30 seconds to segment a 3D model due to the high cost of SAM2’s image encoding at 1024 resolution. Our method, though slightly slower than PartField and Find3D, still belongs to the top-performing group in terms of runtime.

\begin{figure}[H]
  \centering
  \includegraphics[
    width=\columnwidth,
  ]{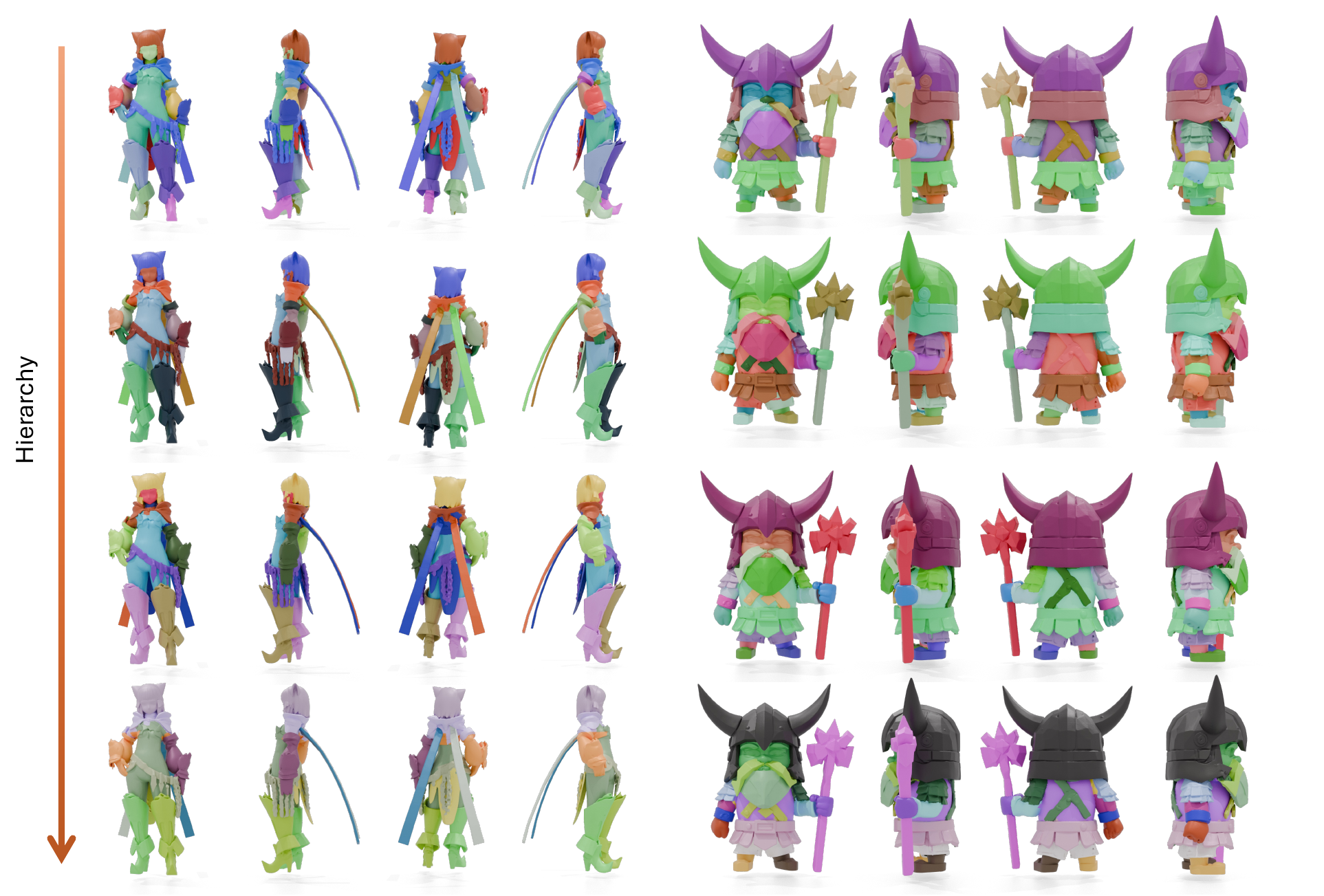}
  \caption{Visualization of hierarchical segmentation results of generative models. 3D models are segmented hierarchically, starting with the finest level of detail at the top, and then proceeding with varying granularities of segmentation for the hands, legs, and head. More results can be found in supplementary material.}
  \label{fig:hierarchy}
\vspace{-10pt}
\end{figure}

\begin{figure}[H]
  \centering
  \includegraphics[width=\linewidth]{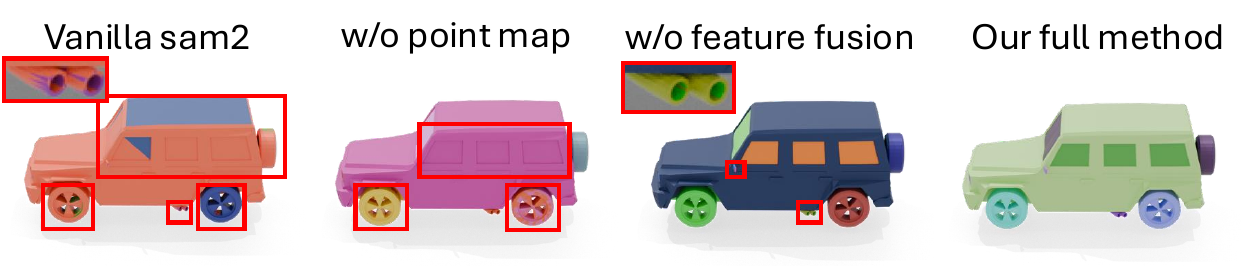}
  \caption{Visualization of ablation study on PartObjaverse-Tiny.
    Anomalies marked in red; subtle defects magnified in upper-left inset.}. 
  \label{fig:ablation}
  \vspace{-2em}
\end{figure}

\begin{figure}[H]
  \centering
  \includegraphics[width=0.8\linewidth]{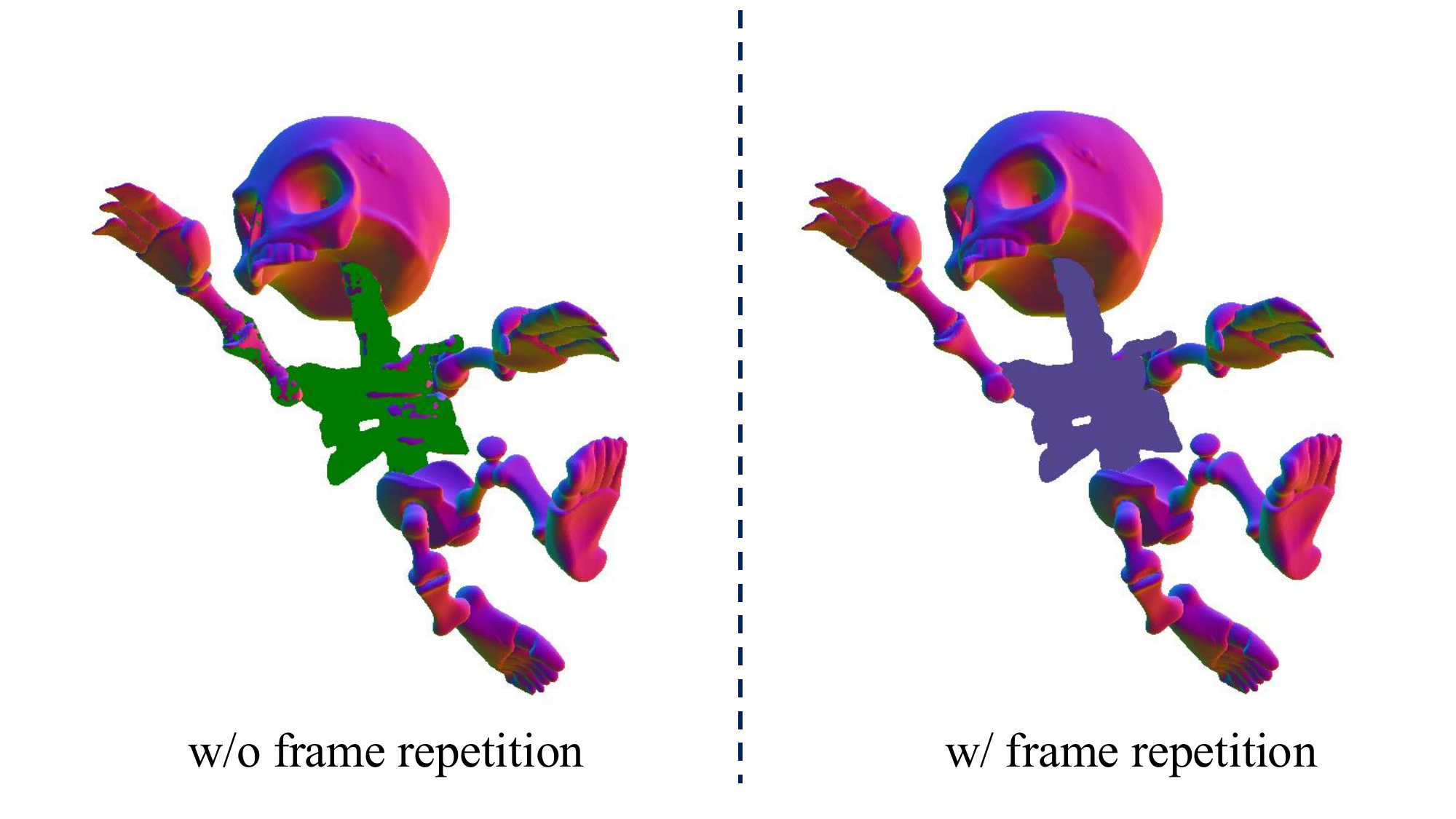}
  \caption{Ablation study of frame repetition}. 
  \label{fig:duplication}
  \vspace{-2em}
\end{figure}

% As post-processing is optional and not central to our approach, its time is not included.

\subsection{Generalization to Generative Model}

% \begin{figure*}
%   \centering
%   \includegraphics[width=\linewidth]{Images/hierarchy.pdf}
%   \caption{Visualization of hierarchical segmentation results of generative models. 3D models are segmented hierarchically, starting with the finest level of detail at the top, and then proceeding with varying granularities of segmentation for the hands, legs, and head.}
%   \label{fig:hierarchy}
% \end{figure*}

To evaluate the generalizability of our method, we extend our method to the generation results. Because the 3D generation model still can not generate handcrafted-like geometry with sharp edges and reasonable structures, segment those generation results is challenging.

We perform hierarchical segmentation on 3D models generated by TripoSG~\cite{triposg} using 2D point prompts on the front view, and automatically infer segmentation on the back view to complete unlabeled regions. As shown in Figure~\ref{fig:hierarchy}, we obtain multi-scale part masks (e.g., hands, legs, heads) and render the results from four orthogonal views for better visualization. Our method demonstrates strong 3D segmentation ability, maintaining clear part awareness even with blurred geometry boundaries.

\subsection{Ablation Study}

% \begin{figure}
%   \centering
%   \includegraphics[width=\linewidth]{Images/ablation_network.pdf}
%   \caption{\footnotesize Illustrations of w/o feature fusion and our full method in ablation study.}
%   \label{fig:comparison_network}
% \end{figure}

We conducted exhaustive ablation experiments on all critical architectural components. We (1) first evaluate on vanilla SAM2 with multi-view normal maps as input, leveraging its inherent capability to segment multi-view consistent masks, (2) then evaluate fine-tuning the image encoder for normal map (without point map) via LoRA, addressing the inherent domain shift between RGB and surface normal distributions. (3) Additionally, we evaluate adding point map and fine-tuning the image encoder for point map via LoRA, injecting point map via a convolution network with a kernel size of 1 (without feature fusion, detailed in supplementary material). As shown in Table~\ref{tab:comparison_abla} and Figure~\ref{fig:ablation}, directly utilizing SAM2 for 3D object segmentation struggles because of SAM2's inability to precisely track. Finetuning SAM2 only on normal modality can improve performance, confirming that LoRA adaptation effectively bridges the domain gap between photometric and geometric encodings. However, the output exhibits failure modes because tracking is still challenging without spatial information. Incorporating point map and fusing features via a convolution network with a kernel size of 1 can addresses spatial ambiguity, but struggles with fine details. Our full method is able to produce satisfactory results, tracking every mask well and segmenting 3D objects without noise.

Repeating the frame with prompt can effectively enhance the segmentation quality. as shown in Figure~\ref{fig:duplication}, the segmentation result with repeating the first frame is more stable and precise, aligned well with the geometry edge.

\begin{figure}
  \centering
  \includegraphics[width=0.8\linewidth]{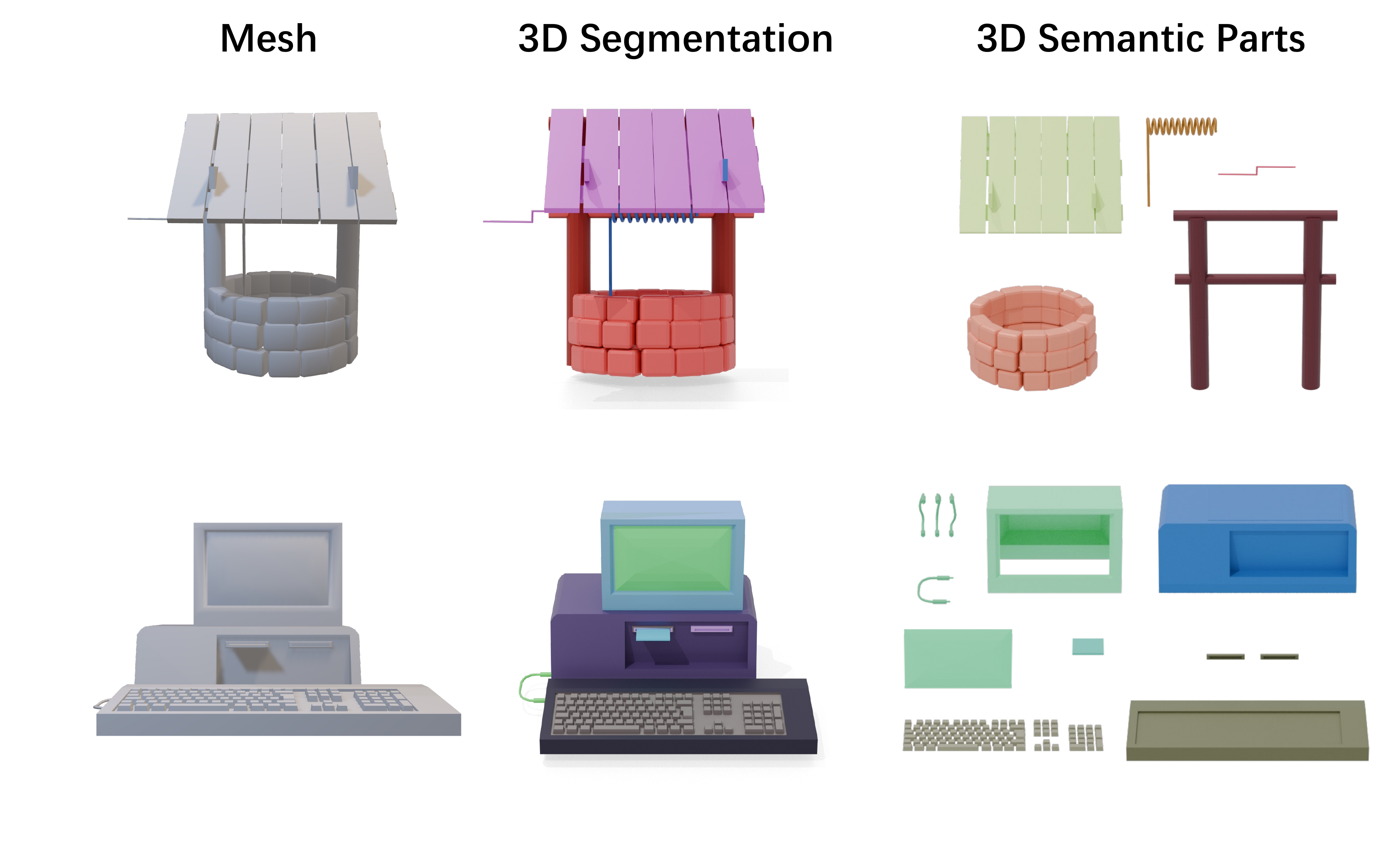}
  \caption{Our method seamlessly integrates with existing zero-shot 3D part completion models, enabling effective zero-shot 3D part amodal segmentation.} % Application. 
  \label{fig:application}
  \vspace{-1.4em}
\end{figure}

\begin{figure}
  \centering
  \includegraphics[width=0.8\linewidth]{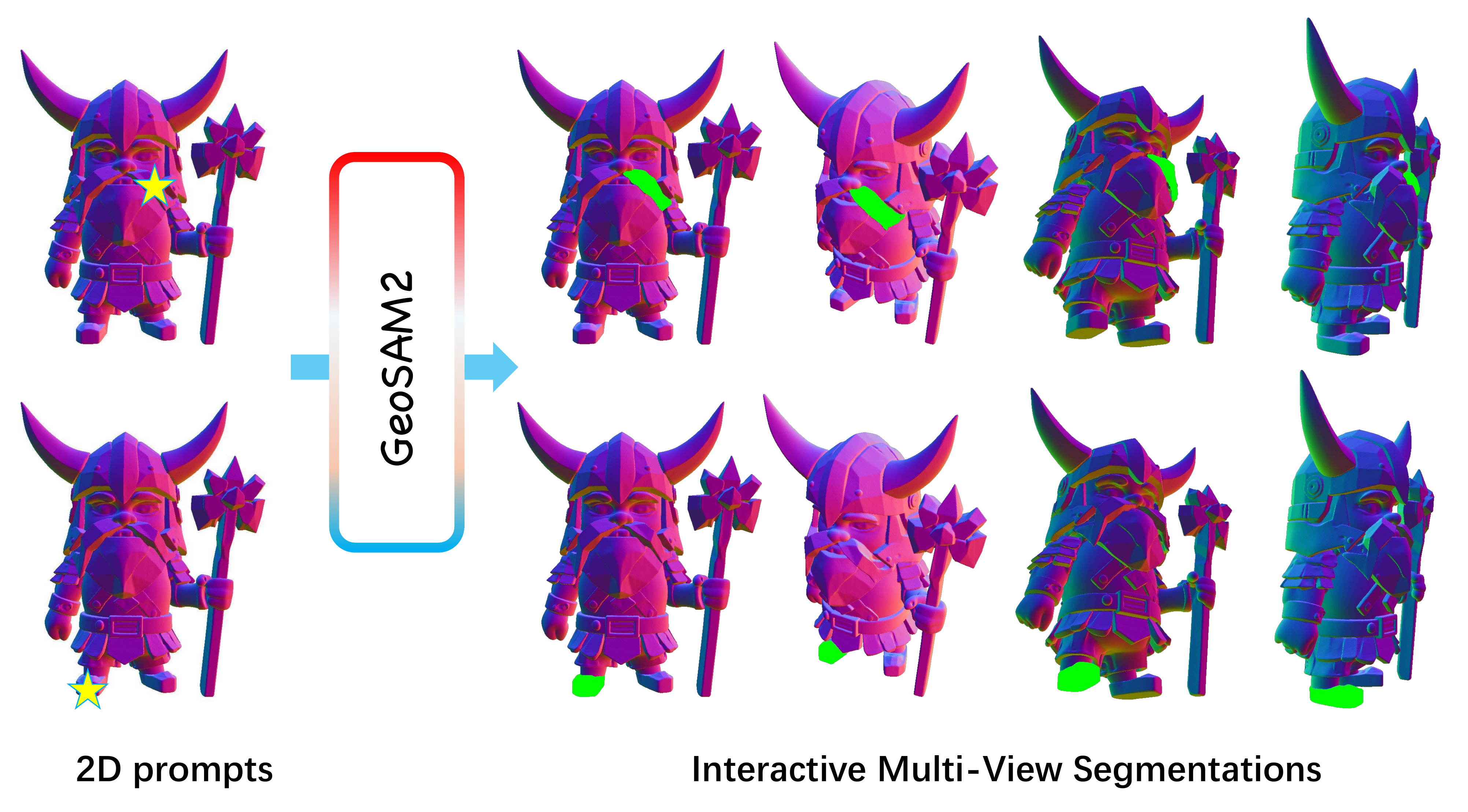}
  \vspace{-2pt}
  \caption{Example of segmenting a 3D part by given a single 2D prompt.} % Application. 
  \label{fig:application_sam}
  \vspace{-15pt}
\end{figure}

\subsection{Application}
The experiment result above shows the strong potential of GeoSAM2 in various applications, including 3D part amodal segmentation, scalpel-grade part segmentation.

\noindent \paragraph{3D part amodal segmentation}
Our method enables precision-controlled segmentation: given a 3D object, users can dynamically adjust part granularity (e.g., separating a chair into legs/back/seat or finer subcomponents) to generate clean, completion-ready segments. This produces artist-grade modular 3D structures where each part maintains watertight boundaries, bridging the gap between generative 3D models and manual modeling workflows. Results can be found in Figure~\ref{fig:application}.

% \noindent \paragraph{scalpel-grade part segmentation}
% Our scalpel-grade segmentation enables precision editing of 3D parts with minimal 2D prompts—extract or merge components at any granularity without iterative scale/text tuning. Need to isolate a human leg or a patch of beard? A single click in 2D propagates to 3D with boundary-exact results, bypassing manual cleanup and preserving structural integrity. Results can be found in Figure~\ref{fig:application_sam}.

\noindent \paragraph{Precise Part Segmentation for 3D Editing} Our segmentation approach enables high-precision 3D part editing with sparse 2D inputs, reducing the need for iterative scale or text tuning. Users can extract or merge components at fine granularity—such as isolating a human leg or a patch of beard—with a single 2D click, which propagates to 3D with well-defined boundaries. The method minimizes manual cleanup while maintaining structural coherence, as demonstrated in Figure~\ref{fig:application_sam}.

% \noindent \paragraph{Resolving 3D Ambiguity via 2D-Guided Editing}
% Traditional graphics workflows involved tedious manual editing of materials/textures or ill-posed inverse rendering optimizations (slow and ambiguous) versus fast but low-quality feedforward methods. Our method leverages 2D prompts to generate 3D masks, enabling intuitive segmentation where material parameters or UV tiling can be directly manipulated in 2D space, transforming ill-posed 3D optimization into well-posed 2D editing. This aligns with conventional modeling pipelines while eliminating ambiguity through human prompts and simplifies geometric edits by bypassing crude bounding boxes in favor of precise region-based control.

\section{Discussion}

\subsection{Limitation}

While our method leverages the tracking ability of SAM2 and is fine-tuned for efficient 2D-controlled 3D segmentation, it still struggles with heavily occluded objects due to its inherently multi-view nature. A promising direction to address this limitation is to incorporate a 3D-aware semantic completion model that further exploits SAM2's priors.

\subsection{Conclusion}
We presented GeoSAM2, a prompt-controllable framework for 3D part segmentation that bridges 2D segmentation priors and 3D geometric reasoning. By casting the task as a multi-view mask prediction problem, our method leverages 2D interaction and pretrained vision models to deliver fine-grained, part-specific segmentation on textureless meshes. Through LoRA-based tuning and residual fusion of geometric features, GeoSAM2 adapts effectively to 3D inputs while preserving the inductive biases of its 2D backbone. Our approach enables intuitive and spatially grounded control via simple prompts—without reliance on per-shape optimization, textual input, or exhaustive 3D annotations. GeoSAM2 paves the way for efficient, accurate, and user-driven 3D segmentation in real-world applications.

\section{Supplementary Material}

In supplmentary material, we first further provide implementation details. We also provide additional discussion on ablation study. Finally, we provide additional visual results. We encourage readers to view our accompanying videos, which showcase the rotation of objects rendered with different colors (each color represents different parts) as presented in the paper.

\subsection{Implementation Details}

\subsubsection{Lift Multi-View Masks to 3D} Given multi-view masks $\{M_i\}_{i=1}^N$ with corresponding camera pose $\{P_i\}_{i=1}^N$ and target mesh $\mathcal{M}$ (normalized to $[-0.5, 0.5]^3$), where $N = 12$, we first randomly sample 5 points on each face of mesh $\mathcal{M}$, and project them to each mask, obtaining depth map $\{D_i\}_{i=1}^N$. To verify the visibility of each point on each, we compare the depth map $\{D_i\}_{i=1}^N$ with the rendered depth map $\{D_ri\}_{i=1}^N$, then adopt those points with depth error $|{D_i} - {D_ri}|_{i=1}^N$ less than 0.001, which precisely reflect the visibility. After obtaining the labels of each point on masks $\{M_i\}_{i=1}^N$, we select the maximum label from those visible multi-view labels of points to get the point-level label.

\subsubsection{Labeling Training Data}

\begin{figure}
  \centering
  \includegraphics[width=\linewidth]{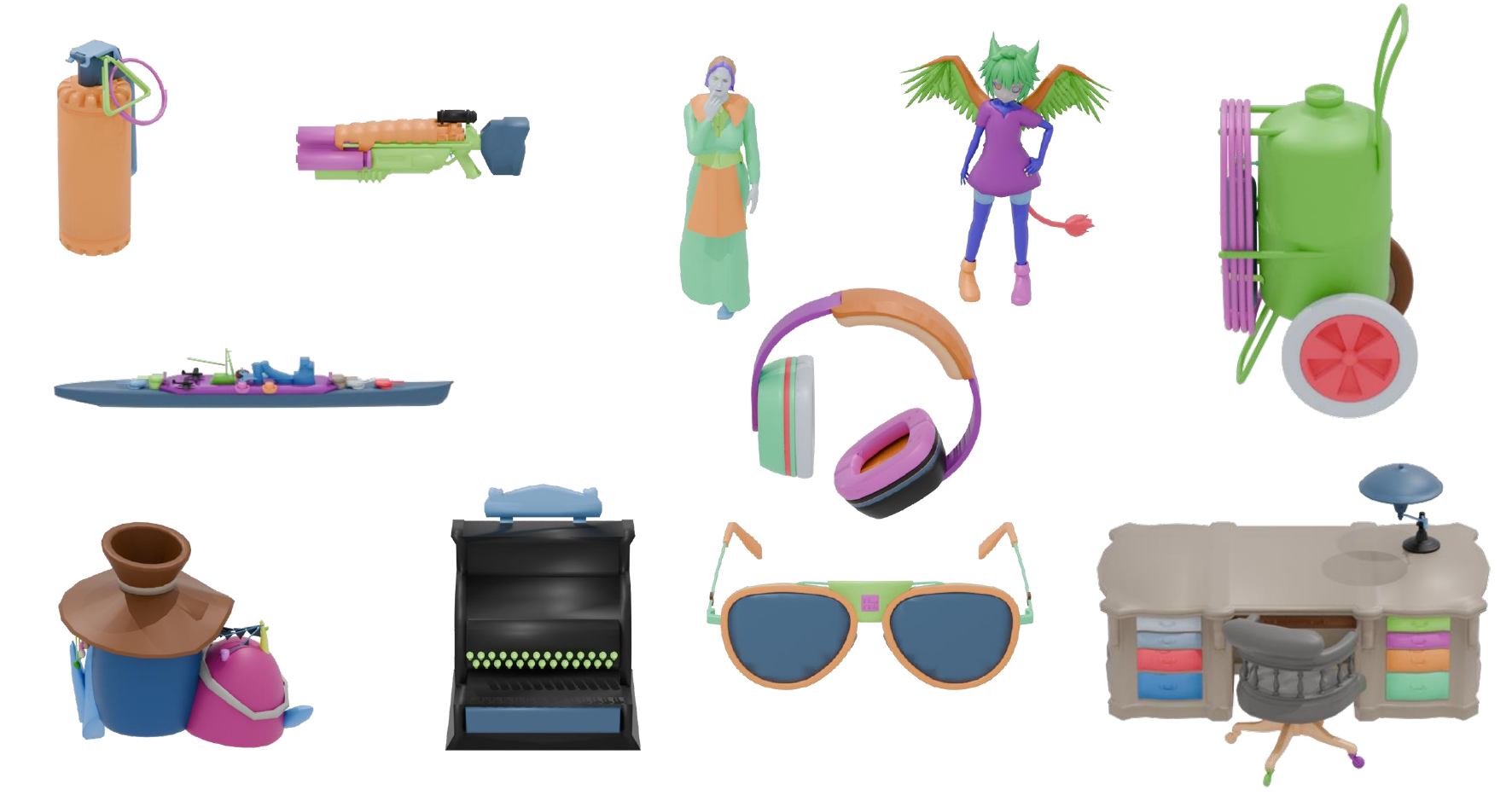}
  \caption{Examples of training data. Each color represents a different part.} 
  \label{fig:trainingdata}
\end{figure}

Our method leverages a diverse collection of open-world 3D data, enriched with fine-grained part annotations for each object. The dataset spans a wide range of object categories and part types, enabling comprehensive evaluation across varied structures. Visualization examples are shown in Figure~\ref{fig:trainingdata}.

Since the 3D models created by artists are often modeled part-by-part—for example, the seat and legs of a chair are modeled as separate components—we first decompose each mesh into parts based on its original geometric connectivity. However, due to varying modeling quality, some models are overly fragmented into many small pieces. To ensure that each part carries meaningful semantic information, we further merge overly fine-grained components into coherent, semantically rich parts.

\subsubsection{Class Grouping of PartNetE} To ensure consistency with previous methods, we group the 45 categories in PartNetE following the categorization protocol used in prior work. The specific category groupings are as follows.

\begin{itemize}
    \item \textbf{Electronics \& Computing Devices:} Keyboard, Mouse, Laptop, Phone, Camera, USB, Display (monitor), Remote, Printer, Switch (if treated as a network or power switch)
    
    \item \textbf{Large Home Appliances:} Washing Machine, Dishwasher, Refrigerator, Oven, Microwave
    
    \item \textbf{Kitchen \& Food-Related Items:} KitchenPot, Kettle, Toaster, CoffeeMachine, Faucet, Dispenser, Knife, Bottle, Bucket (often used in kitchen/cleaning contexts)
    
    \item \textbf{Furniture \& Household Infrastructure:} Table, Chair, FoldingChair, StorageFurniture, Door, Window, Lamp, TrashCan, Safe (often a household or office fixture)
    
    \item \textbf{Tools, Office Supplies, \& Miscellaneous:} Stapler, Scissors, Pen, Pliers, Lighter, Box, Cart (e.g., utility cart), Globe (decorative/educational), Suitcase (travel/personal), Eyeglasses (personal), Clock
\end{itemize}

\subsection{Additional Ablation Study}

% \subsubsection{Repeting Frame}

% Repeating the frame with prompt can effectively enhance the segmentation quality. as shown in Table~\ref{fig:duplication}, the segmentation result with repeating the first frame is more stable and precise, aligned well with the geometry edge.

% \begin{figure}
%   \centering
%   \includegraphics[width=\linewidth]{Images/vis_duplicate.pdf}
%   \caption{Ablation study of frame repetition}. 
%   \label{fig:duplication}
% \end{figure}

\subsubsection{Per-Category Results for Ablation Study}

Due to space constraints in the main paper, we only reported the average performance across categories in our ablation study on different network designs. Here, we provide the full per-category results on PartObjaverse-Tiny and PartNetE to offer a more comprehensive view of the model’s behavior across different object types, illustrated in~\ref{tab:ablation_pot} and~\ref{tab:ablation_partnete}.

% ~\cite{yang2024sampart3d}

\begin{table}[t]
  \centering
  \captionsetup{skip=2pt}
  \caption{\footnotesize Quantitative evaluation of class‑agnostic part segmentation on PartObjaverse-Tiny for ablation study. Instance‑level labels; (mean IoU, \%).}

  \label{tab:ablation_pot}

  \resizebox{\linewidth}{!}{%
  \setlength{\tabcolsep}{2.3pt}
  \begin{tabular}{l|cccccccc|cc}
    \toprule
    \textbf{Method} & Human & Animals & Daily & Build.\& & Transp. & Plants & Food & Elec. & Avg.  \\
    \midrule

    Vanilla SAM2  & 67.04 & 64.34 & 64.37 & 54.88 & 52.05 & 75.78 & 67.16 & 65.46 & 62.59 &  \\
    w/o point map                       & 81.17 & 83.87 & 77.68 & 67.24 & 63.85 & 81.66 & 81.91 & 78.08 & 75.56 &  \\
    w/o feature fusion & 87.27 & 90.87 & 79.98 & 70.81 & 74.75 & 87.64 & \textbf{83.19} & 82.20 & 81.39 &  \\
    \textbf{Ours}                       & \textbf{88.99} & \textbf{91.50} & \textbf{86.04} & \textbf{74.57} & \textbf{77.60} & \textbf{88.92} & 82.72 & \textbf{84.95} & \textbf{84.06} &  \\
    \bottomrule
  \end{tabular}}
\end{table}

% ~\cite{liu2023partsliplowshotsegmentation3d}

\begin{table}[t]
  \centering
  \captionsetup{skip=2pt}
  \caption{\footnotesize Quantitative evaluation of class‑agnostic part segmentation on PartNetE for abltion study. Instance‑level labels; (mean IoU, \%).}
  \label{tab:ablation_partnete}

  \resizebox{\linewidth}{!}{%
  \setlength{\tabcolsep}{2.3pt}
  \begin{tabular}{l|ccccc|c}
    \toprule
    \textbf{Method} & Electro. \& Comput. & Home Appl. & Kitchen \& Food & Furnit. \& Househo. & Tools, Offi., \& Misc. & Avg. \\
    \midrule

    Vanilla SAM2        & 64.33 & 69.26 & 71.73 & 65.10 & 65.41 & 66.55 \\
    w/o point map       & 68.47 & 69.11 & 77.74 & 68.74 & 73.19 & 71.26 \\
    w/o feature fusion  & \textbf{70.18} & \textbf{71.29} & 78.00 & 68.08 & 75.05 & 72.25 \\
    \textbf{Ours}       & 69.93 & 71.23 & \textbf{79.73} & \textbf{71.97} & \textbf{79.41} & \textbf{74.42} \\
    \bottomrule
  \end{tabular}}
\end{table}

\subsubsection{Ablation Study on Feature Fusion} 
\begin{figure}
  \centering
  \includegraphics[width=\linewidth]{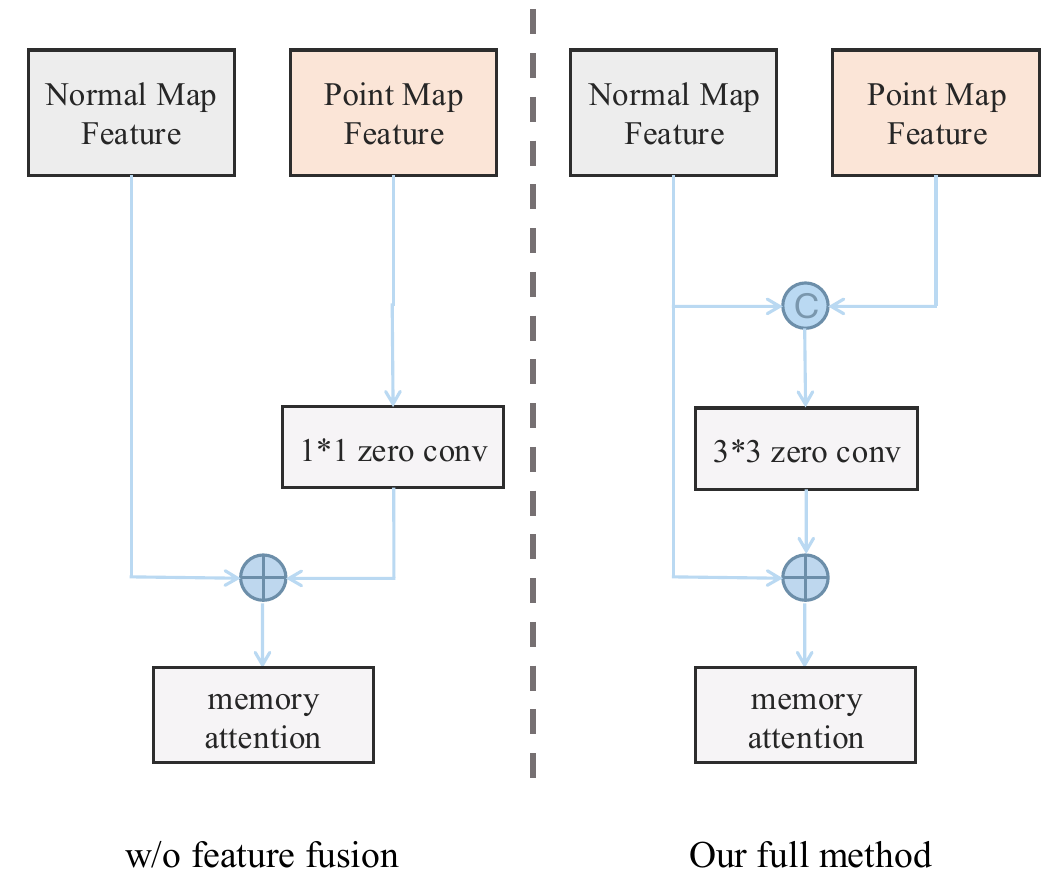}
  \caption{\footnotesize Illustrations of w/o feature fusion and our full method in ablation study.}
  \label{fig:comparison_network}
\end{figure}

To provide a clearer understanding of the design choices in our ablation study on the feature fusion module, we include detailed visualizations of the different network variants used. Figure~\ref{fig:comparison_network} highlight the architectural differences between each setting, helping to better illustrate how various fusion strategies affect performance.

\subsection{Additional Visualization}

\subsubsection{Additional Comparison Visualization}
\begin{figure*}
  \centering
  \includegraphics[page=1,width=0.95\linewidth]{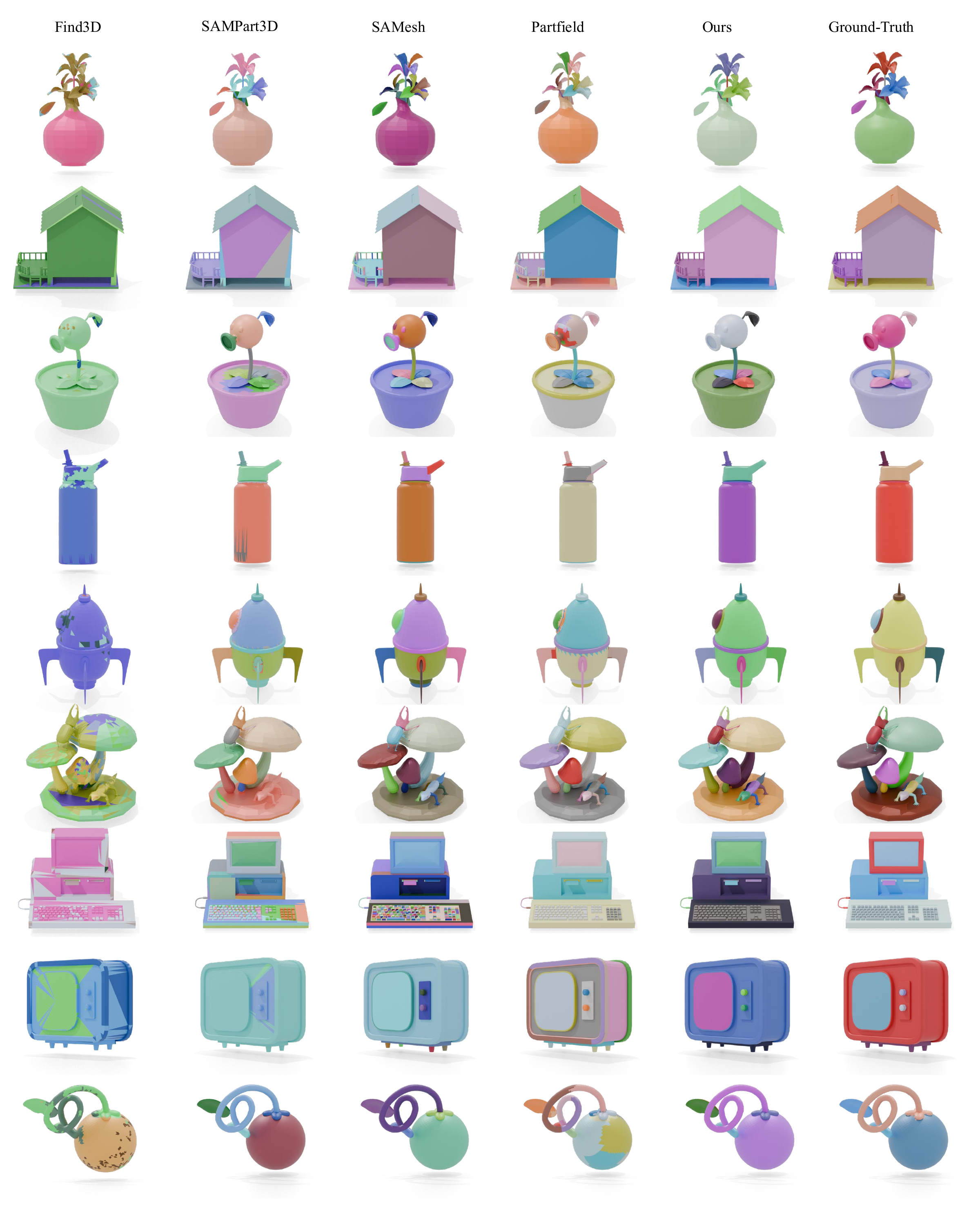}
  \caption{More comparison results on PartObjaverse-Tiny}. 
  \label{fig:compare_pot_supp_1}
\end{figure*}

\begin{figure*}
  \centering
  \includegraphics[width=\linewidth]{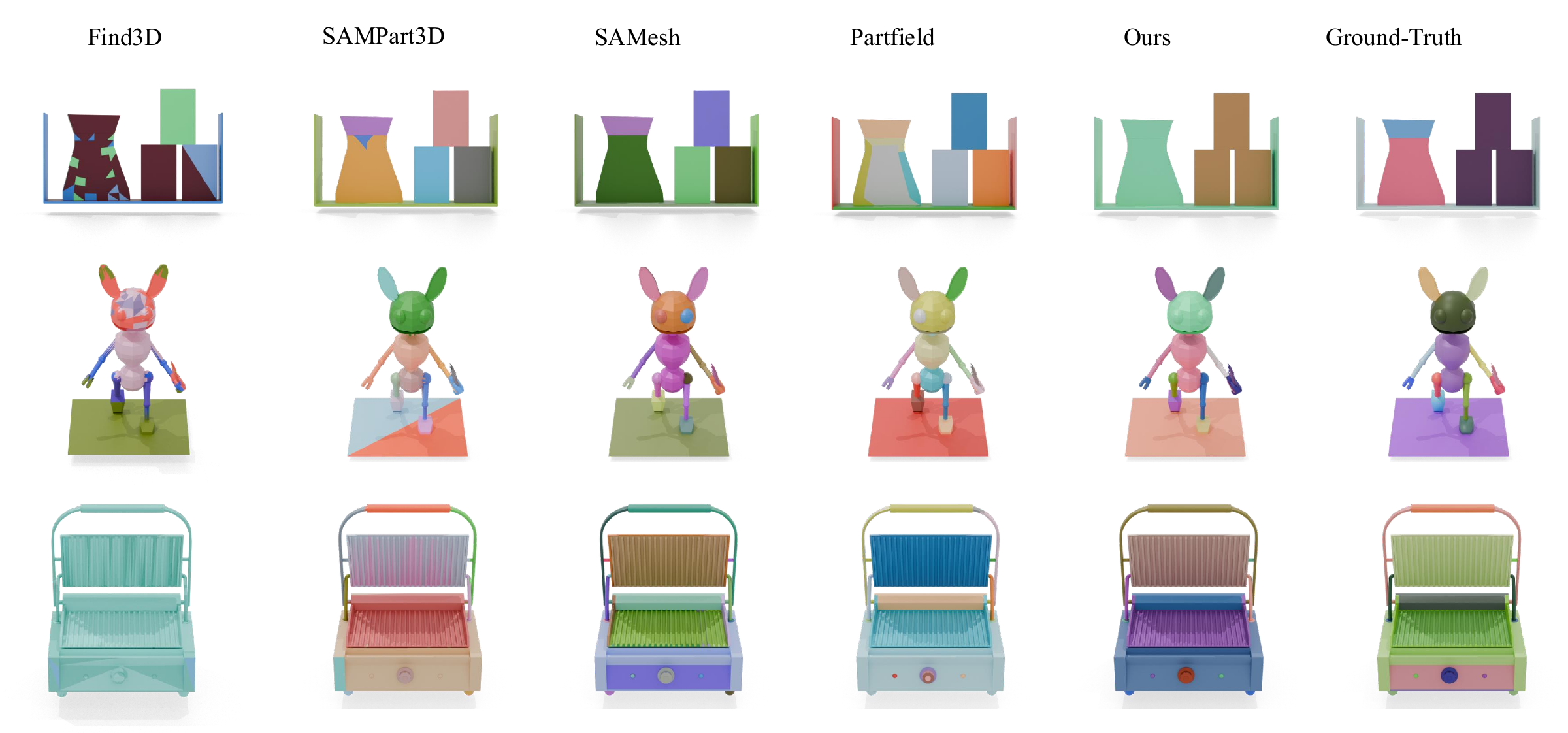}
  \caption{More comparison results on PartObjaverse-Tiny}. 
  \label{fig:compare_pot_supp_2}
\end{figure*}

To further illustrate the superiority of our approach and its robust 3D segmentation performance, we present additional qualitative comparisons on PartObjaverse-Tiny, as shown in Figure~\ref{fig:compare_pot_supp_1} and Figure~\ref{fig:compare_pot_supp_2}.

% \begin{figure*}
%   \centering
%   \includegraphics[width=\linewidth]{Images/hierarchy.pdf}
%   \caption{Visualization of hierarchical segmentation results of generative models. 3D models are segmented hierarchically, starting with the finest level of detail at the top, and then proceeding with varying granularities of segmentation for the hands, legs, and head.}
%   \label{fig:hierarchy}
% \end{figure*}

\subsubsection{Additional Hierachical segmentation Visualization}
\begin{figure*}
  \centering
  \includegraphics[page=1,width=0.9\linewidth]{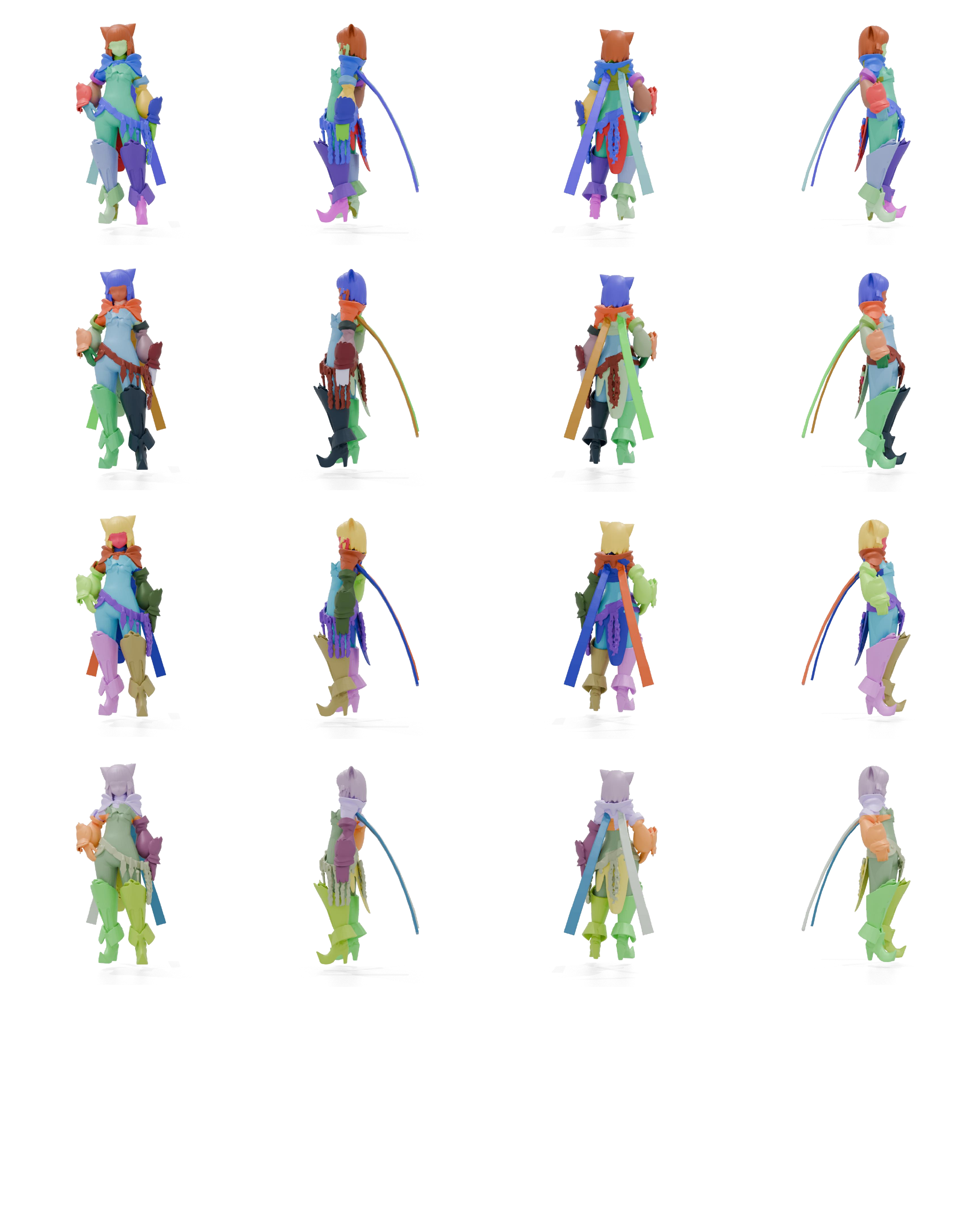}
  \caption{More hierarchical segmentation visualization. 3D models are segmented hierarchically, starting with the finest level of detail at the top, and then proceeding with varying granularities of segmentation for the hands, legs, and head.}. 
  \label{fig:hierarchy_supp_1}
\end{figure*}

\begin{figure*}
  \centering
  \includegraphics[page=2,width=0.9\linewidth]{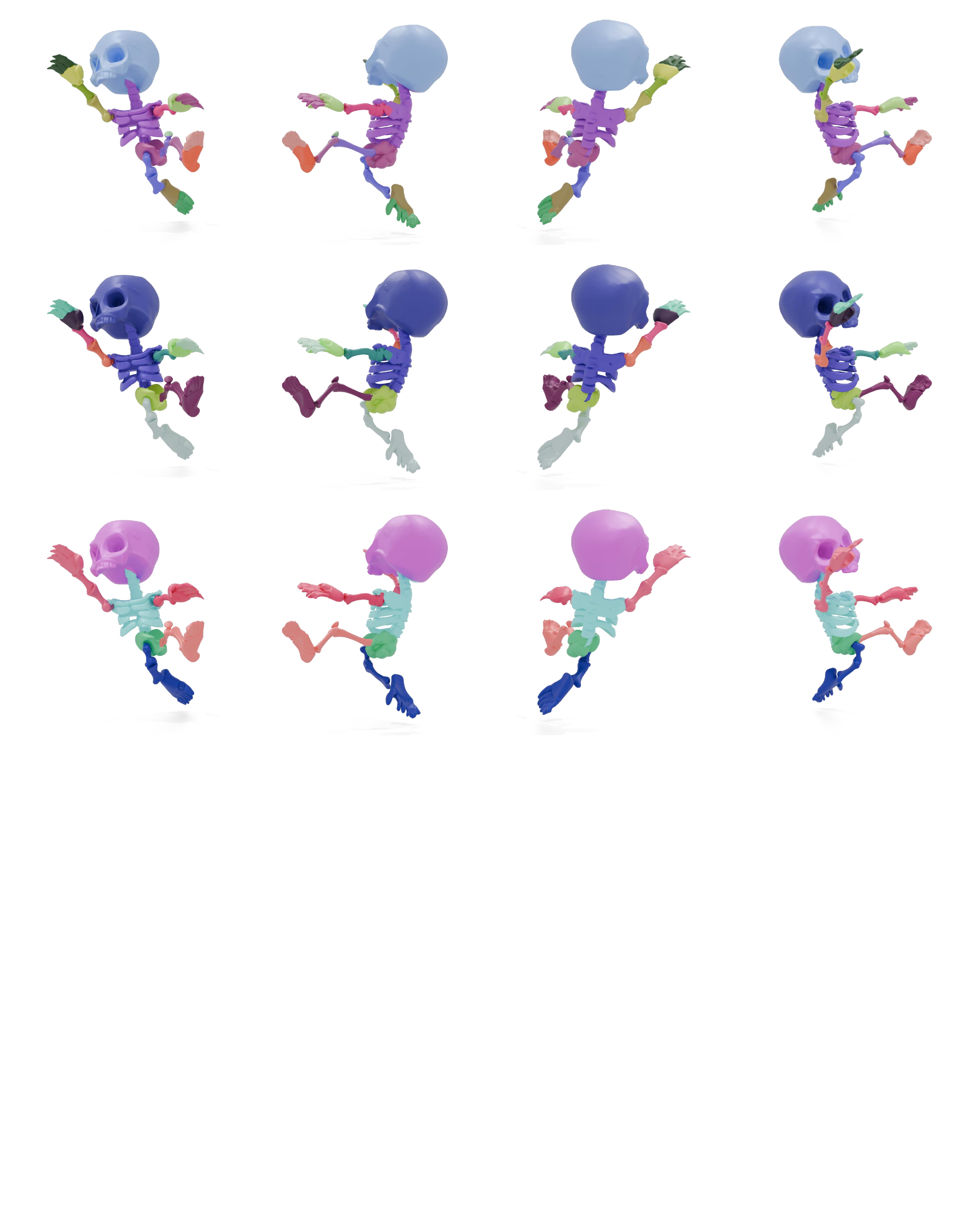}
  \caption{More hierarchical segmentation visualization. 3D models are segmented hierarchically, starting with the finest level of detail at the top, and then proceeding with varying granularities of segmentation for the hands, legs, and head.}. 
  \label{fig:hierarchy_supp_2}
\end{figure*}

\begin{figure*}
  \centering
  \includegraphics[page=3,width=0.9\linewidth]{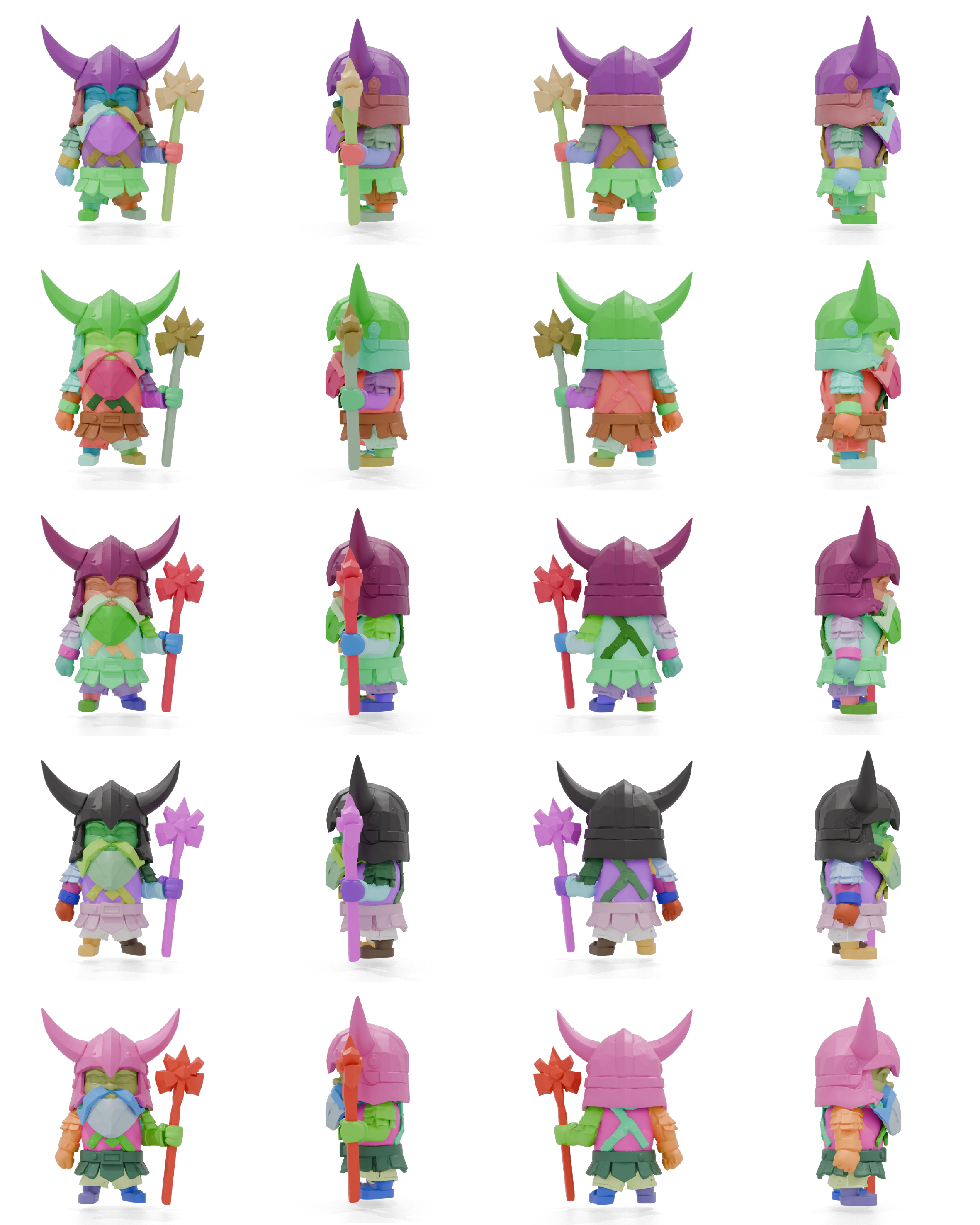}
  \caption{More hierarchical segmentation visualization. 3D models are segmented hierarchically, starting with the finest level of detail at the top, and then proceeding with varying granularities of segmentation for the hands, legs, and head.}. 
  \label{fig:hierarchy_supp_3}
\end{figure*}

We include additional qualitative examples of our hierarchical segmentation results in this subsection, as shown in Figure~\ref{fig:hierarchy_supp_1}, Figure~\ref{fig:hierarchy_supp_2}, and Figure~\ref{fig:hierarchy_supp_3}. These visualizations further demonstrate the effectiveness of our method in producing semantically meaningful and consistent part hierarchies across diverse 3D objects. The results are rendered from multiple viewpoints to better illustrate the multi-level structure and fine-grained part delineation.

% \bibliographystyle{ACM-Reference-Format}
% \bibliography{main}
% \input{sec/append}

% \end{document}
\clearpage
{
    \small
    \bibliographystyle{ieeenat_fullname}
    \bibliography{main}
}

\end{document}